\documentclass[english]{article}
\usepackage{amsfonts}
\usepackage{amsmath}
\usepackage{amssymb}
\usepackage{amsthm}
\usepackage{verbatim}
\usepackage{float}
\usepackage[colorlinks,citecolor=blue,urlcolor=blue,linkcolor=red]{hyperref}
\usepackage{fullpage}
\usepackage[affil-it]{authblk}
\usepackage{siunitx}
\usepackage{textcomp}
\usepackage{tabularx}

\usepackage{graphicx}

\renewcommand{\phi}{\varphi}
\renewcommand{\epsilon}{\varepsilon}

\begin{document}
\title{A Deep Learning Approach to Denoise Optical Coherence Tomography Images of the Optic Nerve Head}

\author[1 $\dagger$]{Sripad Krishna Devalla}
\author[1 $\dagger$]{Giridhar Subramanian}
\author[1,2]{Tan Hung Pham}
\author[3,1]{Xiaofei Wang}
\author[2,4]{Shamira Perera}
\author[2,1]{Tin A. Tun}
\author[2,4]{Tin Aung}
\author[2,4,5,6,7]{Leopold Schmetterer}
\author[5 $\star$]{Alexandre H. Thi\'{e}ry}
\author[1,3 $\star$]{Micha\"el J. A. Girard}

\affil[1]{Ophthalmic Engineering and Innovation Laboratory, Department of Biomedical Engineering, Faculty of Engineering, National University of Singapore, Singapore.}
\affil[2]{Singapore Eye Research Institute, Singapore National Eye Centre, Singapore.}
\affil[3]{Beijing Advanced Innovation Center for Biomedical Engineering, School of Biological Science and Medical Engineering, Beihang University, Beijing, China.}
\affil[4]{Duke-NUS Graduate Medical School.}
\affil[5]{Department of Statistics and Applied Probability, National University of Singapore, Singapore.}
\affil[6]{Nanyang Technological University, Singapore.}
\affil[7]{Department of Clinical Pharmacology, Medical University of Vienna, Austria.}
\affil[8]{Center for Medical Physics and Biomedical Engineering, Medical University of Vienna, Austria.}
\bigskip
\affil[$\dagger$]{Both authors contributed equally and are both first authors.}
\affil[$\star$]{Both authors contributed equally and are both corresponding authors.}

\maketitle


%
%
\begin{abstract}
\noindent
\\
\textbf{Purpose.}
To develop a deep learning approach to de-noise optical coherence tomography (OCT) B-scans of the optic nerve head (ONH). 
\\
\\
\textbf{Methods.}
Volume scans consisting of 97 horizontal B-scans were acquired through the center of the ONH using a commercial OCT device (Spectralis) for both eyes of 20 subjects. For each eye, single-frame (without signal averaging), and multi-frame (75x signal averaging) volume scans were obtained. A custom deep learning network was then designed and trained with 2,328 ‘clean B-scans’(multi-frame B-scans), and their corresponding ‘noisy B-scans’(clean B-scans + gaussian noise) to de-noise the single-frame B-scans. The performance of the de-noising algorithm was assessed qualitatively, and quantitatively on 1,552 B-scans using the signal to noise ratio (SNR), contrast to noise ratio (CNR), and mean structural similarity index metrics (MSSIM).
\\
\\
\textbf{Results.}
The proposed algorithm successfully denoised unseen single-frame OCT B-scans. The denoised B-scans were qualitatively similar to their corresponding multi-frame B-scans, with enhanced visibility of the ONH tissues. The mean SNR  increased from $4.02 \pm 0.68$ dB (single-frame) to $8.14 \pm 1.03$ dB (denoised). For all the ONH tissues, the mean CNR increased from $3.50 \pm 0.56$ (single-frame) to $7.63 \pm 1.81$ (denoised). The MSSIM increased from $0.13 \pm 0.02$ (single frame) to $0.65 \pm 0.03$ (denoised) when compared with the corresponding multi-frame B-scans.
\\
\\
\textbf{Conclusions.}
Our deep learning algorithm can denoise a single-frame OCT B-scan of the ONH in under 20 ms, thus offering a framework to obtain superior quality OCT B-scans with reduced scanning times and minimal patient discomfort.

\end{abstract}

%
%
\section{Introduction}
\label{sec.intro}
In recent years, optical coherence tomography (OCT) imaging has become a well-established clinical tool for assessing optic nerve head (ONH) tissues, and for monitoring many ocular \cite{RN62,RN98} and neuro-ocular pathologies \cite{RN96}. However, despite several advancements in OCT technology \cite{RN41}, the quality of B-scans is still hampered by speckle noise \cite{RN4,RN5,RN6,RN7,RN16,RN17,RN18}, low signal strength \cite{RN8}, blink \cite{RN8,RN59} and motion artefacts \cite{RN8,RN20}.\\

Specifically, the granular pattern of speckle noise deteriorates the image contrast, making it difficult to resolve small and low-intensity structures (e.g., sub-retinal layers) \cite{RN4,RN5,RN6}, thus affecting the clinical interpretation of OCT data. Also, poor image contrast can lead to automated segmentation errors \cite{RN21,RN58,RN19}, and incorrect tissue thickness estimation \cite{RN60}, potentially affecting clinical decisions. For instance, segmentation errors for the retinal nerve fiber layer (RNFL) thickness can lead to over/under estimation of glaucoma \cite{RN133}.\\

Currently, there exist many hardware \cite{RN159,RN156,RN155,RN157,RN158,RN24,RN25,RN154,RN151} and software schemes \cite{RN151,RN100,RN10} to de-noise OCT B-scans. Hardware approaches offer robust noise suppression through frequency compounding \cite{RN24,RN25,RN154,RN151} and multi-frame averaging (spatial compounding) \cite{RN159,RN156,RN155,RN157,RN158}. While multi-frame averaging techniques have shown to enhance image quality and presentation \cite{RN100,RN10}, they are sensitive to registration errors \cite{RN10}, and require longer scanning times \cite{RN11}. Moreover, elderly patients often face discomfort and strain \cite{RN1}, when they remain fixated for long durations \cite{RN1,RN3}. Software techniques, on the other hand, attempt to denoise through numerical algorithms \cite{RN4,RN5,RN6,RN7,RN16,RN17,RN18} or filtering techniques \cite{RN161,RN162,RN160}. However, registration errors \cite{RN37}, computational complexity \cite{RN4,RN31,RN30,RN64}, and sensitivity to choice of parameters \cite{RN35} limit their usage in the clinic.\\

In this study, we propose a deep learning approach to denoise OCT B-scans. We aimed to obtain multi-frame quality B-scans (i.e. signal-averaged) from single-frame (without signal averaging) B-scans of the ONH. We hope to offer a denoising framework to obtain superior quality B-scans, with reduced scanning duration and minimal patient discomfort.

\section{Methods}
\subsection{Patient Recruitment}
A total of 20 healthy subjects were recruited at the Singapore National Eye Centre. All subjects gave written informed consent. This study adhered to the tenets of the Declaration of Helsinki and was approved by the institutional review board of the hospital. The inclusion criteria for healthy subjects were: an intraocular pressure (IOP) less than 21 mmHg, and healthy optic nerves with a vertical cup-disc ratio (VCDR) less than or equal to 0.5.\\

\subsection{Optical Coherence Tomography Imaging}

The subjects were seated and imaged under dark room conditions by a single operator (TAT). A spectral-domain OCT (Spectralis, Heidelberg Engineering, Heidelberg, Germany) was used to image both eyes of each subject. Each OCT volume consisted of 97 horizontal B-scans ($32-\mu$m distance between B-scans; 384 A-scans per B-scan), covering a rectangular area of 15\textdegree x 10\textdegree centered on the ONH. For each eye, single-frame (without signal averaging), and multi-frame (75x signal averaging) volume scans were obtained. Enhanced depth imaging (EDI) \cite{RN61} and eye tracking \cite{RN12,RN70} modalities were used during the acquisition. From all the subjects, we obtained a total of 3,880 B-scans for each type of scan (single-frame or multi-frame).\\

\subsection{Volume Registration}
The multi-frame volumes were reoriented to align with the single-frame volumes through rigid translation/rotation transformations using 3D software (Amira, version 5.6; FEI). This registration was performed using a voxel-based algorithm that maximized mutual information between two volumes \cite{RN72}. Registration was essential to quantitatively validate the corresponding regions between the denoised and multi-frame B-scans. Note that Spectralis follow-up mode was not used in this study. Although the follow-up mode allows a new scanning of the same area by identifying previous scan locations, in many cases, it can distort B-scans and thus provide unrealistic tissue structures in the new scan. \\

\subsection{Deep Learning Based Denoising}
While deep learning has shown promising segmentation \cite{RN107, RN104, RN106, RN105}, classification \cite{RN108, RN109, RN110}, and denoising \cite{RN113, RN111, RN112} applications in the field of medical imaging for modalities such as magnetic resonance imaging (MRI), its application to OCT imaging is still in its infancy \cite{RN120, RN119, RN118, RN117, RN116, RN115, RN84, RN103, RN82, RN123, RN122, RN121}. Although recent deep learning studies have shown successful segmentation \cite{RN120, RN119, RN106, RN117, RN116, RN115, RN84, RN103, RN82} and classification applications \cite{RN123, RN122,RN121} in OCT imaging, to the best of our knowledge no study exists yet to assess the success of denoising OCT B-scans. We believe, a denoising framework would not only increase the reliability of clinical information in single-frame B-scans, but also improve the robustness of segmentation and classification tools. \\

In this study, we developed a fully-convolutional neural network, inspired by our earlier DRUNET architecture \cite{RN103} to denoise single-frame OCT B-scans of the ONH. It leverages on the inherent advantages of U-Net \cite{RN74}, residual learning \cite{RN76}, dilated convolutions \cite{RN77},  and multi-scale hierarchical feature extraction \cite{RN90} to obtain multi-frame quality B-scans. Briefly, the U-Net and its skip connections helped the network learn both the local (tissue texture) and contextual information (spatial arrangement of tissues). The contextual information was further exploited using dilated convolution filters. Residual connections improved the flow of the gradient information through the network, and multi-scale hierarchical feature extraction helped restore tissue boundaries in the B-scans. \\

\subsection{Network Architecture}

\begin{figure}[H]
    \centering
    \includegraphics[width=.80\textwidth]{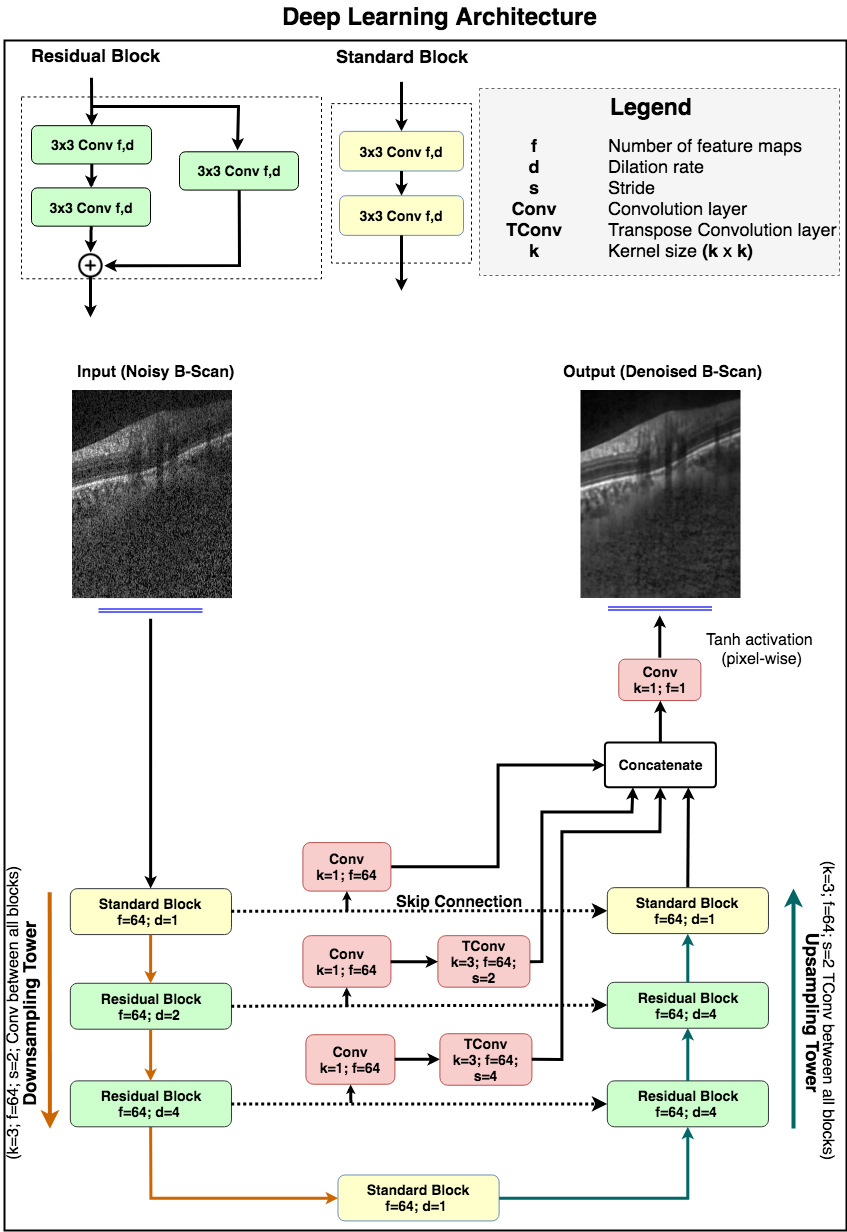}
    \caption{The architecture comprised of two towers: \textbf{(1)} A downsampling tower – to capture the contextual information (i.e., spatial arrangement of the tissues), and \textbf{(2)} an upsampling tower – to capture the local information (i.e., tissue texture). Each tower consisted of two blocks: \textbf{(1)} a standard block, and \textbf{(2)} a residual block. The latent space was implemented as a standard block. The multi-scale hierarchical feature extraction unit helped better recover tissue edges eroded by speckle noise. The network consisted of 900k trainable parameters.}
    \label{fig:1}
\end{figure}

The network was composed of a downsampling and an upsampling tower, connected to each other via skip-connections (\textbf{Figure \ref{fig:1}}). Each tower consisted of one standard block and two residual blocks. Both the standard and the residual blocks comprised of two dilated convolution layers (64 filters; size = 3x3). A 3x3 convolution layer was used to implement the identity connection in the residual block. \\

In the downsampling tower, an input B-scan (size: 496x384) was fed to a standard block (dilation rate: 1) followed by two residual blocks (dilation rate: 2 and 4, respectively). A convolution layer (64 filters; size = 3x3; stride = 2) after every block sequentially reduced the dimensionality, enabling the network to understand the contextual information. \\

The latent space was implemented as a standard block (dilation rate: 1) to transfer the feature maps from the downsampling to the upsampling tower. \\

The upsampling tower helped the network capture the local information. It consisted of two residual blocks (dilation rate: 4) and a standard block (dilation rate: 1). After every block, a transpose convolution layer (64 filters; size = 3x3; stride = 2) was used to restore the B-scan sequentially to its original dimension.\\

Multi-scale hierarchical feature extraction \cite{RN90} helped recover tissue boundaries eroded by speckle noise in the single-frame B-scans. It was implemented by passing the feature maps at each downsampling level through a convolution layer (64 filters; size = 1x1), followed by a transpose convolution layer (64 filters; size = 3x3) to restore the original B-scan resolution. The restored maps were then concatenated with the output feature maps from the upsampling tower.\\

Finally, the concatenated feature maps were fed to the output convolution layer (1 filter; size = 1x1), followed by pixel-wise hyperbolic tangent ($\tanh$) activation to produce a denoised OCT B-scan. \\

In both towers, all layers except the last output layer, were activated by an exponential linear unit (ELU) \cite{RN91} function. In addition, in each residual block, the feature maps were batch normalized \cite{RN92} and ELU activated before addition. \\

The proposed network comprised of 900,000 trainable parameters. The network was trained end-to-end using the Adam optimizer \cite{RN93}, and we used the mean absolute error as loss function. We trained and tested the proposed network on an NVIDIA GTX 1080 founders edition GPU with CUDA v8.0 and cuDNN v5.1 acceleration. With the given hardware configuration, each single-frame OCT B-scan was denoised under 20 ms. \\

\subsection{Training and Testing of the Network}

From the dataset of 3,880 B-scans, 2,328 of them (from both eyes of 12 subjects) were used as a part of the training dataset. The training set consisted of  ‘clean’ B-scans and their corresponding ‘noisy’ versions. The ‘clean’ B-scans were simply the multi-frame (75x signal averaging) B-scans. The ‘noisy’ B-scans were generated by adding Gaussian noise ($\mu = 0$ and $\sigma = 1$) to the respective ‘clean’ B-scans (\textbf{Figure \ref{fig:2}}). \\

The testing set consisted of 1,552 single-frame B-scans (from both eyes of 8 subjects) to be denoised. We ensured that the scans from the same subject weren’t used in both training and testing sets. \\

\subsection{Data Augmentation}
An exhaustive offline data augmentation was done to circumvent the scarcity of training data. We used elastic deformations \cite{RN94, RN103}, rotations (clockwise and anti-clockwise; $10^{\circ}$), occluding patches \cite{RN103}, and horizontal flipping for both ‘clean’ and ‘noisy’ B-scans. Briefly, elastic deformations were used to produce the combined effects of shearing and stretching in an attempt to make the network invariant to atypical morphologies (as seen in glaucoma \cite{RN95}). Ten occluding patches of size 60 x 20 pixels were added at random locations to non-linearly reduce (pixel intensities multiplied by a random factor between 0.2 and 0.8) the visibility of the ONH tissues. This was done to make the network invariant to blood vessel shadows that are common in OCT B-scans \cite{RN163}. Note that a full description of our data augmentation approach can be found in our previous paper \cite{RN103}. \\

\begin{figure}[H]
    \centering
    \includegraphics[width=1.0\textwidth]{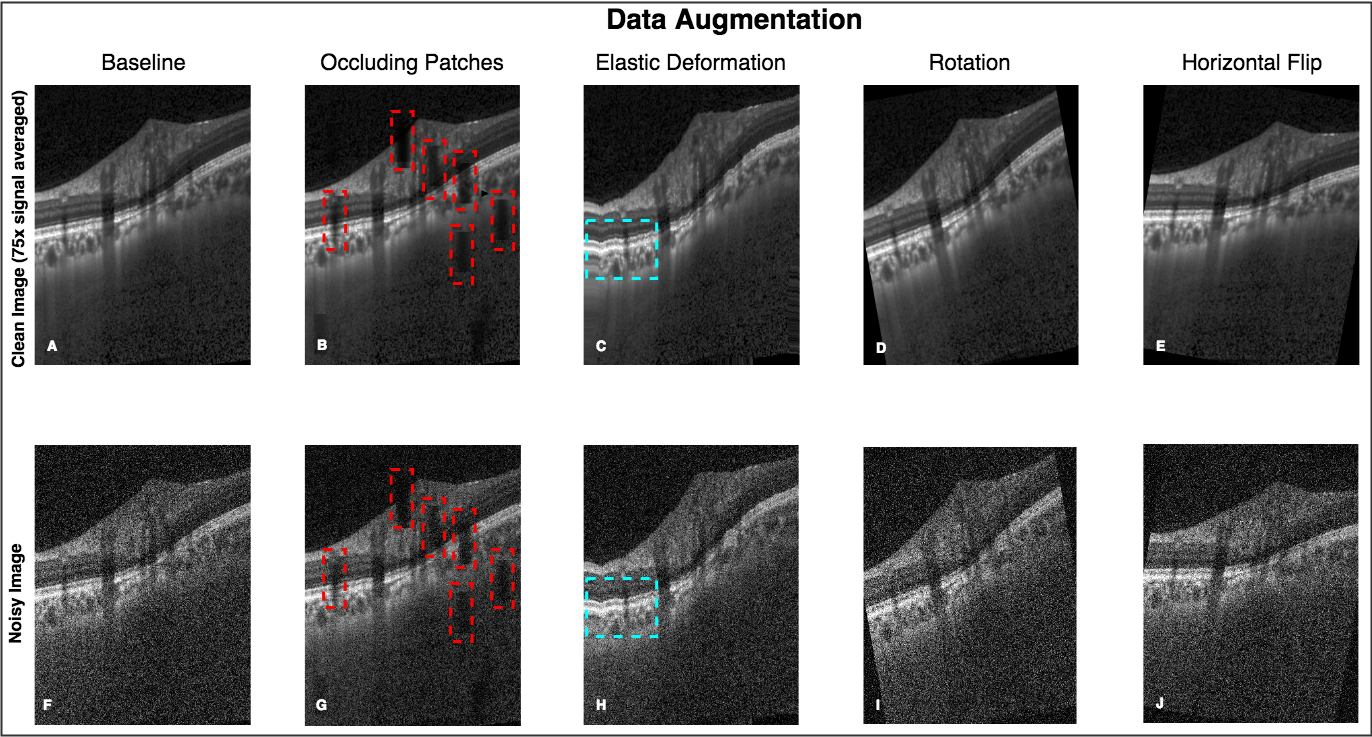}
    \caption{An exhaustive offline data augmentation was done to circumvent the scarcity of training data. \textbf{(A-E)} represent the original and the data augmented ‘clean’ B-scans (multi-frame). \textbf{(F-J)} represent the same for the corresponding ‘noisy’ B-scans. The occluding patches \textbf{(B and G; red boxes)} were added to make the network robust in the presence of blood vessel shadows. Elastic deformations \textbf{(C and H; cyan boxes)} were used to make the network invariant to atypical morphologies. A total of 23,280 B-scans of each type (clean/noisy) were generated from 2,328 baseline B-scans.}
    \label{fig:2}
\end{figure}

Using data augmentation, we were able to generate a total of 23,280 ‘clean’ and 23,280 corresponding ‘noisy’ B-scans that were added to the training dataset. An example of data augmentation performed on a single ‘clean’ and corresponding ‘noisy’ B-scan is shown in (\textbf{Figure \ref{fig:2}}).\\

\subsection{Denoising Performance – Qualitative Analysis}

All denoised single-frame B-scans were manually reviewed by expert observers (SD \& GS) and qualitatively compared against their corresponding multi-frame B-scans. \\

\subsection{Denoising Performance – Quantitative Analysis}

The following image quality metrics were used to assess the denoising performance of the proposed algorithm: \textbf{(1)} signal to noise ratio (SNR); \textbf{(2)} contrast to noise ratio (CNR); and \textbf{(3)} mean structural similarity index measure (MSSIM) \cite{RN166}. These metrics were computed for the single-frame, multi-frame, and denoised OCT B-scans (all from the testing set; 1,552 B-scans of each type). \\

The SNR (expressed in dB) was a measure of signal strength relative to noise. It was defined as: \\

\begin{align*}
\textrm{SNR}_i = -10 \times \log_{10}\left( \frac{\|f_o - \widetilde{f}\|^2}{\|f_o\|^2} \right)
\end{align*}

where $f_o$ is the ‘clean’(multi-frame) B-scan, and $\widetilde{f}$ the B-scan to be compared with $f_o$(either the ‘noisy’ [single frame] or the denoised B-scan). A high SNR value indicates low noise in the given B-scan with respect to the ‘clean’ B-scan.\\

The CNR was a measure of contrast difference between different tissue layers. It was defined as: \\

\begin{align*}
\textrm{CNR}_i = \frac{|\mu_r-\mu_b|}{\sqrt{0.5(\sigma_b^2+\sigma_b^2)}}
\end{align*}

where $\mu_r$ and $\sigma_r^2$ denoted the mean and variance of pixel intensity for a chosen ROI within the tissue ‘i’ in a given B-scan, while $\mu_b$ and $\sigma_b^2$ represented the same for the background ROI. The background ROI was chosen as a 20 x 384 (in pixels) region at the top of the image (within the vitreous). A high CNR value suggested enhanced visibility of the given tissue. \\

The CNR was computed for the following tissues: \textbf{(1)} RNFL; \textbf{(2)} ganglion cell layer + inner plexiform layer (GCL + IPL); \textbf{(3)} all other retinal layers; \textbf{(4)} retinal pigment epithelium (RPE); \textbf{(5)} peripapillary choroid; \textbf{(6)} peripapillary sclera; and \textbf{(7)} lamina cribrosa (LC). Note that the CNR was computed only in the visible portions of the peripapillary sclera and LC. For each tissue, the CNR was computed as the mean of twenty five ROIs (8x8 pixels each) in a given B-scan. \\

The structural similarity index measure (SSIM) \cite{RN166} was computed to assess the changes in tissue structures (i.e., edges) between the single-frame/denoised B-scans and the corresponding multi-frame B-scans (ground-truth). The SSIM was defined between -1 and +1, where -1 represented ‘no similarity’, and +1 ‘perfect similarity’. It was defined as:\\

\begin{align*}
\textrm{SSIM}(x,y) = \frac{(2\mu_x\mu_y+C_1)(2\sigma_{xy}+C_2)}{(\mu_x^2+\mu_y^2+C_1)(\sigma_x^2+\sigma_y^2+C_2)}
\end{align*}

where x and y represented the denoised and multi-frame B-scan respectively; $\mu_x$, $\sigma_x^2$ denoted the mean intensity and standard deviation of the chosen ROI in B-scan x, while $\mu_y$, $\sigma_y^2$ represented the same for B-scan y; $\sigma_{xy}$ represented the cross-covariance of the ROIs in B-scans x and y. C1 and C2  (constants to stabilize the division) were chosen as 6.50 and 58.52, as recommended in a previous study \cite{RN166}.\\

The MSSIM was computed as the mean of SSIM from  ROIs (8x8 pixels each) across an B-scan (stride=1; scanned horizontally). It was defined as: \\

\begin{align*}
\textrm{MSSIM}(X,Y) =\frac{1}{M} \sum_{k=1}^{M} \textrm{SSIM}(x_k,y_k)
\end{align*}

Note that the SNR, and MSSIM were computed for an entire B-scan, as opposed to the CNR that was computed for individual tissues.\\

\section{Results}
\subsection{Qualitative Analysis}
\begin{figure}[H]
    \centering
    \includegraphics[width=1.0\textwidth]{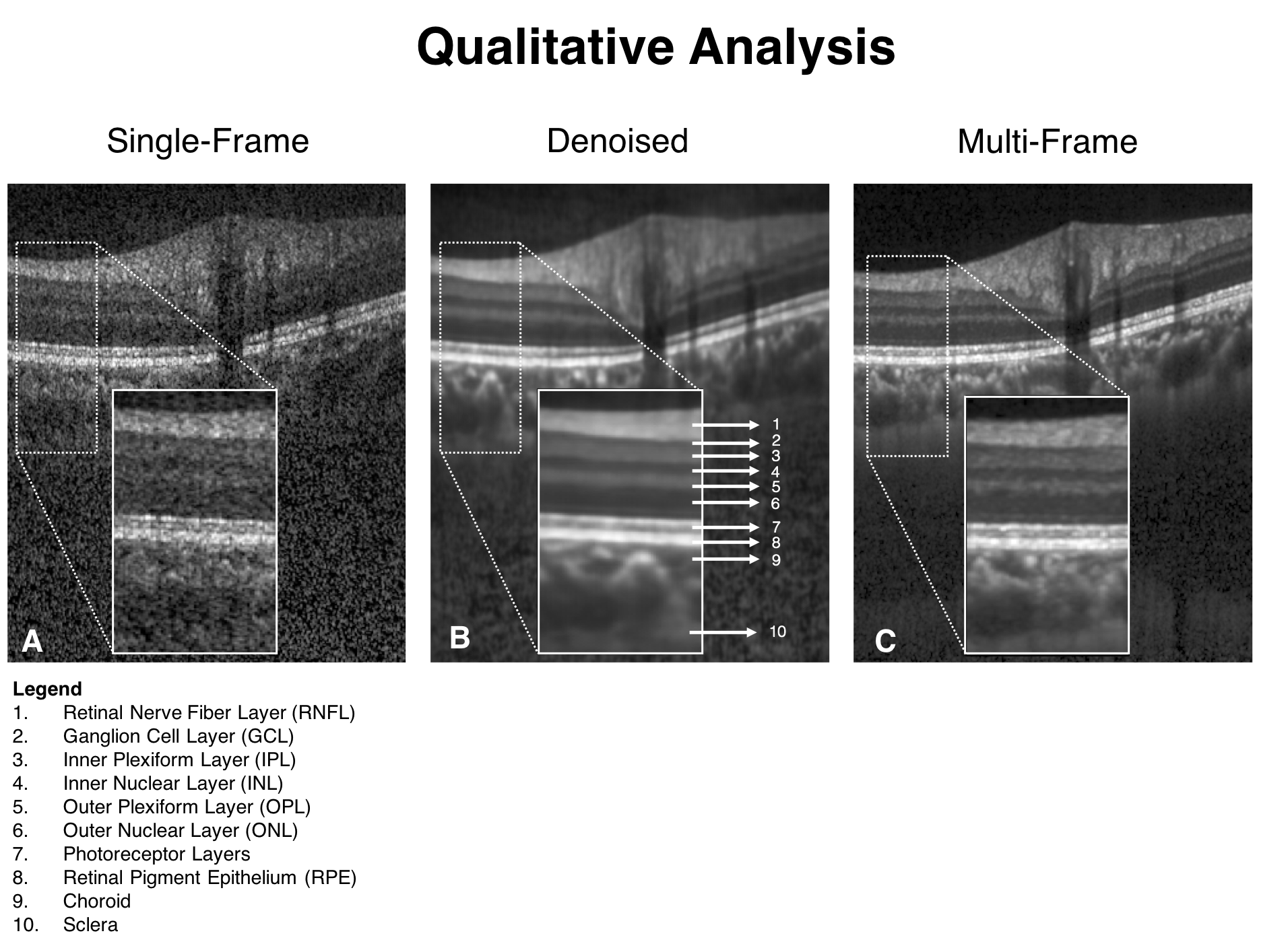}
    \caption{Single-frame \textbf{(A)}, denoised \textbf{(B)}, and multi-frame \textbf{(C)} B-scans for a healthy subject are shown. The denoised B-scan can be observed to be qualitatively similar to its corresponding multi-frame B-scan. Specifically, the visibility of the retinal layers, and choroid, and lamina cribrosa were prominently improved. Sharp and clear boundaries were also obtained for retinal layers, and the choroid-scleral interface.}
    \label{fig:3}
\end{figure}

\begin{figure}[H]
    \centering
    \includegraphics[width=0.65\textwidth]{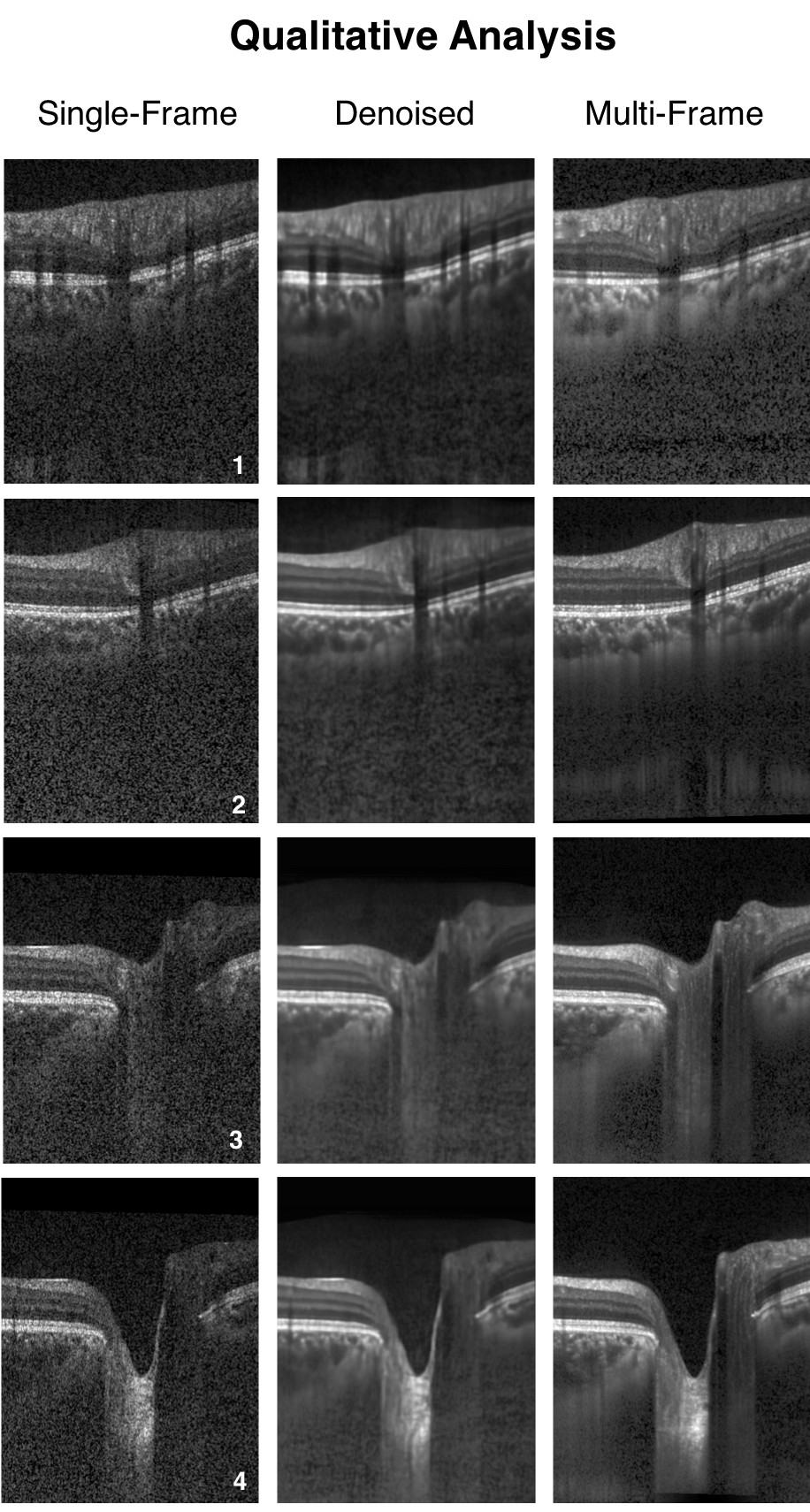}
    \caption{Single-frame, denoised and multi-frame B-scans for four healthy subjects \textbf{(1-4)} are shown. In all cases, the denoised B-scans \textbf{(2nd column)} were consistently similar (qualitatively) to their corresponding multi-frame B-scans \textbf{(3rd column)}.}
    \label{fig:4}
\end{figure}

When trained with the ‘clean’ B-scans (multi-frame) and the corresponding ‘noisy’ B-scans, our network was able to successfully denoise unseen single-frame B-scans. The single-frame, denoised and multi-frame B-scan for a healthy subject can be found in (\textbf{Figure \ref{fig:3}}). In all the cases, the denoised B-scans were qualitatively similar to their corresponding multi-frame B-scans (\textbf{Figure \ref{fig:4}}). Overall, the visibility of all ONH tissues were prominently enhanced (\textbf{Figure \ref{fig:3}; B}). \\


\subsection{Quantitative Analysis}
On average, we observed a two-fold increase in SNR upon denoising. Specifically, the mean SNR for the unseen single-frame/denoised B-scans were: $4.02 \pm 0.68$ dB / $8.14 \pm 1.03$ dB respectively, when computed against their respective multi-frame B-scans.\\

In all cases, the multi-frame B-scans always offered a higher CNR compared to their corresponding single-frame B-scans. Further, the denoised B-scans consistently offered a higher CNR compared to the single-frame B-scans, for all tissues. Specifically, the mean CNR (Table 1) for the for the single-frame/denoised/multi-frame B-scans were:$2.97 \pm 0.42/7.28 \pm 0.63/5.18 \pm 0.76$ for the RNFL, $3.83 \pm 0.43/12.09 \pm 4.22/11.62 \pm 1.85$ for the GCL + IPL, $2.71 \pm 0.33/5.61 \pm 1.46/4.62 \pm 0.86 $for all other retinal layers, $5.62 \pm 0.72/9.25 \pm 2.25/8.10 \pm 1.44$ for the RPE, $2.99 \pm 0.43/5.99 \pm 0.45/5.75 \pm 0.63$ for the choroid, $2.42 \pm 0.39/6.40 \pm 1.68/6.00 \pm 0.96$ for the sclera, and $4.02 \pm 1.23/6.81 \pm 1.99/6.46 \pm 1.81$ for the LC. \\

On average, our denoising approach offered a five-fold increase in MSSIM. Specifically, the mean MSSIM for the single-frame/denoised B-scans were: $0.13 \pm 0.02/ 0.65 \pm 0.03$, when computed against their respective multi-frame B-scans. \\

\section{Discussion}
In this study, we present a custom deep learning approach to denoise single-frame OCT B-scans of the ONH. When trained with the ‘clean’ (multi-frame) and the corresponding  ‘noisy’ B-scans, our network denoised unseen single-frame B-scans.  The proposed network leveraged on the inherent advantages of U-Net, residual learning, and dilated convolutions \cite{RN103}. Further, the multi-scale hierarchical feature extraction \cite{RN90} pathway helped the network recover ONH tissue boundaries degraded by speckle noise. Having successfully trained, tested and validated our network on 1,552 single-frame OCT B-scans of the ONH, we observed a consistently higher SNR and CNR for all ONH tissues, and a consistent five-fold increase in the MSSIM in all the denoised B-scans. Thus, we may be able to offer a robust deep learning framework to obtain superior quality OCT B-scans with reduced scanning duration and minimal patient discomfort. \\

Using the proposed network, we obtained denoised B-scans that were qualitatively similar to their corresponding multi-frame B-scans (\textbf{Figure \ref{fig:3}}) and (\textbf{Figure \ref{fig:4}}), owing to the reduction in noise levels. The mean SNR for the denoised B-scans was $8.14 \pm 1.03$ dB, a two-fold improvement (reduction in noise level) from improvement from $4.02 \pm 0.68$ dB that was obtained for the single-frame B-scans, thus offering an enhanced visibility of the ONH tissues. Given the significance of the neural (retinal layers) \cite{RN130,RN45,RN129,RN46,RN47} and connective tissues (sclera and LC) \cite{RN56,RN57,RN52,RN54,RN53}, in ocular pathologies such as glaucoma \cite{RN98}, and age-related macular degeneration \cite{RN127}, their enhanced visibility is critical in a clinical setting. Furthermore, reduced noise levels would likely increase the robustness of aligning/registration algorithms used to monitor structural changes over time \cite{RN60}. This is crucial for the management of multiple ocular pathologies \cite{RN131,RN132}. \\

In denoised B-scans (vs single-frame B-scans), we consistently observed higher contrast across tissues. Our approach enhanced the visibility of small (e.g. RPE and photoreceptors) and low-intensity tissues (e.g. GCL and IPL; \textbf{Figure \ref{fig:3} B}). For all tissues, the mean CNR increased from $3.50 \pm 0.56$ (single-frame) to $7.63 \pm 1.81$ (denoised). Since existing automated segmentation algorithms rely on high contrast, we believe that our approach could potentially reduce the likelihood of segmentation errors that are relatively common in commercial algorithms \cite{RN21,RN58,RN19,RN22}. For instance, the incorrect segmentation of the RNFL can lead to inaccurate thickness measurements, leading to under-/over- estimation of glaucoma \cite{RN133}. By using the denoising framework as a precursor to automated segmentation/thickness measurement, we could increase the reliability \cite{RN288} of such clinical tools. \\

Upon denoising, we observed a five-fold increase in MSSIM (single-frame/denoised:  $0.13 \pm 0.02/ 0.65 \pm 0.03$), when validated against the multi-frame B-scans. The preservation of features and structural information plays an important role in accurately measuring cellular level disruption to determine retinal pathology.  For instance, the measurement of the ellipsoid zone (EZ) disruption \cite{RN136} provides an insight into the photoreceptor structure, that is significant in pathologies such as diabetic retinopathy \cite{RN139}, macular hole \cite{RN137}, macular degeneration \cite{RN138}, and ocular trauma \cite{RN140}. Existing multi-frame averaging techniques \cite{RN10} significantly enhance and preserve the integrity of the structural information by supressing speckle noise \cite{RN11,RN64,RN35,RN61}. However, they are limited by a major clinical challenge: the inability of the patients to remain fixated for long scanning times,\cite{RN1,RN3} and the resultant discomfort \cite{RN1}.\\

In this study, we are proposing a methodology to significantly reduce scanning time while enhancing OCT signal quality. In our healthy subjects, it took on average 3.5 min to capture a ‘clean’ (multi frame) volume, and 25 s for a ‘noisy’ (single frame) volume. Since we can denoise a single B-scan in 20 ms (or 2 s for a volume of 97 B-scans), this means that we can theoretically generate a denoised OCT volume in about 27 seconds (= time of acquisition of the ‘noisy’ volume [25 s] + denoising processing [2 s]). Thus, we may be able to drastically reduce the scanning duration by more than 7 folds, while maintaining superior image quality. \\

Besides speckle noise, patient dependent factors such as cataract \cite{RN143, RN144, RN142,RN145} and/ or lack of tear film in dry eye can significantly diminish OCT scan quality \cite{RN8,RN143, RN144, RN142,RN145,RN141}. While lubricating eye drops and frequent blinking can instantly improve image quality for patients with corneal drying \cite{RN141,RN146}, the detrimental effects of cataract on OCT image quality might be reduced only if cataract surgery is performed \cite{RN8,RN143,RN144}. Moreover, pupillary dilation might be needed especially in subjects with small pupil sizes to obtain acceptable quality B-scans \cite{RN8,RN147}, which is highly crucial in the monitoring of glaucoma \cite{RN147}. Pupillary dilation is also time consuming and may cause patient discomfort \cite{RN148}.  It is plausible that the proposed framework, when extended, could be a solution to the afore-mentioned factors that limit image quality, avoiding the need for any additional clinical procedure.\\

In this study, several limitations warrant further discussion. First, the proposed network was trained and tested only on B-scans from one device (Spectralis). Every commercial OCT device has its own proprietary algorithm to pre-process the raw OCT data, potentially presenting a noise distribution different from what our network was trained with. Hence, we are unsure of our network’s performance on other devices. Nevertheless, we offer a proof of concept which could be validated by other groups on multiple commercial OCT devices.\\

Second, we were unable to train our network with a speckle noise model representative of the Spectralis device. Such a model is currently not provided by the manufacturer and would be extremely hard to reverse-engineer because information about all pre- and post-processing done to the OCT signal is also not provided. While there exist a number of OCT denoising studies that assume a Rayleigh \cite{RN7}/ Generalized Gamma distribution to describe speckle noise \cite{RN64}, we observed that they were ill-suited for our network. From our experiments, the best denoising performance was obtained when our network was trained with a simple Gaussian noise model ( $\mu =0$ and $\sigma = 1$). It is possible that a thorough understanding of the raw noise distribution prior to the custom pre-processing on the OCT device could improve the performance of our network. We aim to test this hypothesis with a custom-built OCT system in our future works. \\

Third, while we have discussed the need for reliable clinical information from poor quality OCT scans, that could be critical for the diagnosis and management of ocular pathology (e.g., glaucoma) , we have yet to test the networks’ performance on pathological B-scans. \\

Fourth, we observed that the SNR and CNR metrics were higher for the denoised B-scans than their corresponding multi-frame B-scans. This could be attributed to over-smoothening (or blurring) of tissue textures that was consistently present in the denoised B-scans. We are currently exploring other deep learning techniques to improve the B-scan sharpness that is lost during denoising. \\

Fifth, we were unable to provide further validation of our algorithm by comparing our outputs to histology data. Such a validation would be extremely difficult, as one would need to first image a human ONH with OCT, process with histology, and register both datasets. Furthermore, while we believe our algorithm is able to restore tissue texture accurately (when comparing denoised B-scans with multi-frame B-scans), an exact validation of our approach is not possible. Long  fixation times in obtaining the multi-frame B-scans lead to subtle motion artifacts (eye movements caused by microsaccades or unstable fixation) \cite{RN12}, displaced optic disc center \cite{RN165}, and axial misalignment \cite{RN8}, causing minor registration errors between the single-frame and multi-frame B-scans, thus preventing an exact comparison between the denoised B-scans and the multi-frame B-scans. \\

Finally, no quantitative measurements were performed on the denoised images to assess differences in tissue morphology between the denoised and multi-frame B-scans. Undertaking this work in the future could increase the clinical relevance of the denoised B-scans.\\

In conclusion, we have developed a custom deep learning approach to denoise single-frame OCT B-scans. With the proposed network, we were able to denoise a single-frame OCT B-scan in under 20 ms. We hope that the proposed framework could resolve the current trade-off in obtaining reliable and superior quality scans, with reduced scanning times and minimal patient discomfort. Finally, we believe that our approach may be helpful for low-cost OCT devices, whose noisy B-scans may be enhanced by artificial intelligence (as opposed to expensive hardware) to the same quality as in current commercial devices. \\

\section*{Acknowledgments}
Singapore Ministry of Education Academic Research Funds Tier 1 (R-155-000-168-112 [AHT];R-397-000-294-114 [MJAG]); National University of Singapore (NUS) Young Investigator Award Grant (NUSYIA$\textunderscore$FY16$\textunderscore$P16; R-155-000-180-133; AHT); National University of Singapore Young Investigator Award Grant (NUSYIA$\textunderscore$FY13$\textunderscore$P03; R-397-000-174-133 [MJAG]); Singapore Ministry of Education Academic Research Funds Tier 2 (R-397-000-280-112 [MJAG]);National Medical Research Council (Grant NMRC/STAR/0023/2014 [TA]).\\

\section*{Disclosures}
The authors declare that there are no conflicts of interest related to this article.\\

\bibliographystyle{unsrt}
\bibliography{DenoisingBib}

\begin{thebibliography}{100}

\bibitem{RN62}
Mehreen Adhi and Jay~S. Duker.
\newblock Optical coherence tomography – current and future applications.
\newblock {\em Current opinion in ophthalmology}, 24(3):213--221, 2013.

\bibitem{RN98}
Igor~I. Bussel, Gadi Wollstein, and Joel~S. Schuman.
\newblock Oct for glaucoma diagnosis, screening and detection of glaucoma
  progression.
\newblock {\em British Journal of Ophthalmology}, 98(Suppl 2):ii15, 2014.

\bibitem{RN96}
Ramiro~S. Maldonado, Pradeep Mettu, Mays El-Dairi, and M.~Tariq Bhatti.
\newblock The application of optical coherence tomography in neurologic
  diseases.
\newblock {\em Neurology: Clinical Practice}, 5(5):460--469, 2015.

\bibitem{RN41}
M.~E. van Velthoven, D.~J. Faber, F.~D. Verbraak, T.~G. van Leeuwen, and M.~D.
  de~Smet.
\newblock Recent developments in optical coherence tomography for imaging the
  retina.
\newblock {\em Prog Retin Eye Res}, 26(1):57--77, 2007.

\bibitem{RN4}
Guoying Feng Zhongping~Chen Yongzhao~Du, Gangjun~Liu.
\newblock Speckle reduction in optical coherence tomography images based on
  wave atoms.
\newblock {\em Journal of Biomedical Optics}, 19(5), May 2014.

\bibitem{RN5}
M.~Bashkansky and J.~Reintjes.
\newblock Statistics and reduction of speckle in optical coherence tomography.
\newblock {\em Optics Letters}, 25(8):545--547, 2000.

\bibitem{RN6}
Kin Man~Yung Joseph M.~Schmitt, S. H.~Xiang.
\newblock Speckle in optical coherence tomography.
\newblock {\em Journal of Biomedical Optics}, 4(1), January 1999.

\bibitem{RN7}
Ahmadreza Baghaie, Zeyun Yu, and Roshan~M. D’Souza.
\newblock State-of-the-art in retinal optical coherence tomography image
  analysis.
\newblock {\em Quantitative Imaging in Medicine and Surgery}, 5(4):603--617,
  2015.

\bibitem{RN16}
M.~Szkulmowski, I.~Gorczynska, D.~Szlag, M.~Sylwestrzak, A.~Kowalczyk, and
  M.~Wojtkowski.
\newblock Efficient reduction of speckle noise in optical coherence tomography.
\newblock {\em Opt Express}, 20(2):1337--59, 2012.

\bibitem{RN17}
Mahdad Esmaeili, Alireza~Mehri Dehnavi, Hossein Rabbani, and Fedra Hajizadeh.
\newblock Speckle noise reduction in optical coherence tomography using
  two-dimensional curvelet-based dictionary learning.
\newblock {\em Journal of Medical Signals and Sensors}, 7(2):86--91, 2017.

\bibitem{RN18}
Zhongping Jian, Zhaoxia Yu, Lingfeng Yu, Bin Rao, Zhongping Chen, and Bruce~J.
  Tromberg.
\newblock Speckle attenuation in optical coherence tomography by curvelet
  shrinkage.
\newblock {\em Optics Letters}, 34(10):1516--1518, 2009.

\bibitem{RN8}
Seth C. Nelson Diana~Chao Joshua S.~Hardin, Giovanni~Taibbi and Gianmarco
  Vizzeri.
\newblock Factors affecting cirrus-hd oct optic disc scan quality: A review
  with case examples.
\newblock {\em Journal of Ophthalmology}, 2015, 2015.

\bibitem{RN59}
Kaweh Mansouri, Felipe~A. Medeiros, Andrew~J. Tatham, Nicholas Marchase, and
  Robert~N. Weinreb.
\newblock Evaluation of retinal and choroidal thickness by swept-source optical
  coherence tomography: Repeatability and assessment of artifacts.
\newblock {\em American journal of ophthalmology}, 157(5):1022--1032, 2014.

\bibitem{RN20}
S.~Asrani, L.~Essaid, B.~D. Alder, and C.~Santiago-Turla.
\newblock Artifacts in spectral-domain optical coherence tomography
  measurements in glaucoma.
\newblock {\em JAMA Ophthalmol}, 132(4):396--402, 2014.

\bibitem{RN21}
Y.~Liu, H.~Simavli, C.~J. Que, J.~L. Rizzo, E.~Tsikata, R.~Maurer, and T.~C.
  Chen.
\newblock Patient characteristics associated with artifacts in spectralis
  optical coherence tomography imaging of the retinal nerve fiber layer in
  glaucoma.
\newblock {\em Am J Ophthalmol}, 159(3):565--76.e2, 2015.

\bibitem{RN58}
S.~Asrani, L.~Essaid, B.~D. Alder, and C.~Santiago-Turla.
\newblock Artifacts in spectral-domain optical coherence tomography
  measurements in glaucoma.
\newblock {\em JAMA Ophthalmology}, 132(4):396--402, 2014.

\bibitem{RN19}
K.~E. Kim, J.~W. Jeoung, K.~H. Park, D.~M. Kim, and S.~H. Kim.
\newblock Diagnostic classification of macular ganglion cell and retinal nerve
  fiber layer analysis: differentiation of false-positives from glaucoma.
\newblock {\em Ophthalmology}, 122(3):502--10, 2015.

\bibitem{RN60}
Madhusudhanan Balasubramanian, Christopher Bowd, Gianmarco Vizzeri, Robert~N.
  Weinreb, and Linda~M. Zangwill.
\newblock Effect of image quality on tissue thickness measurements obtained
  with spectral-domain optical coherence tomography.
\newblock {\em Optics express}, 17(5):4019--4036, 2009.

\bibitem{RN133}
S.~L. Mansberger, S.~A. Menda, B.~A. Fortune, S.~K. Gardiner, and S.~Demirel.
\newblock Automated segmentation errors when using optical coherence tomography
  to measure retinal nerve fiber layer thickness in glaucoma.
\newblock {\em Am J Ophthalmol}, 174:1--8, 2017.

\bibitem{RN159}
N.~Iftimia, B.~E. Bouma, and G.~J. Tearney.
\newblock Speckle reduction in optical coherence tomography by "path length
  encoded" angular compounding.
\newblock {\em J Biomed Opt}, 8(2):260--3, 2003.

\bibitem{RN156}
A.~E. Desjardins, B.~J. Vakoc, W.~Y. Oh, S.~M. Motaghiannezam, G.~J. Tearney,
  and B.~E. Bouma.
\newblock Angle-resolved optical coherence tomography with sequential angular
  selectivity for speckle reduction.
\newblock {\em Opt Express}, 15(10):6200--9, 2007.

\bibitem{RN155}
T.~Bajraszewski, M.~Wojtkowski, M.~Szkulmowski, A.~Szkulmowska, R.~Huber, and
  A.~Kowalczyk.
\newblock Improved spectral optical coherence tomography using optical
  frequency comb.
\newblock {\em Opt Express}, 16(6):4163--76, 2008.

\bibitem{RN157}
B.~F. Kennedy, T.~R. Hillman, A.~Curatolo, and D.~D. Sampson.
\newblock Speckle reduction in optical coherence tomography by strain
  compounding.
\newblock {\em Opt Lett}, 35(14):2445--7, 2010.

\bibitem{RN158}
T.~Klein, R.~Andre, W.~Wieser, T.~Pfeiffer, and R.~Huber.
\newblock Joint aperture detection for speckle reduction and increased
  collection efficiency in ophthalmic mhz oct.
\newblock {\em Biomed Opt Express}, 4(4):619--34, 2013.

\bibitem{RN24}
M.~Pircher, E.~Gotzinger, R.~Leitgeb, A.~F. Fercher, and C.~K. Hitzenberger.
\newblock Speckle reduction in optical coherence tomography by frequency
  compounding.
\newblock {\em J Biomed Opt}, 8(3):565--9, 2003.

\bibitem{RN25}
J.~M. Schmitt.
\newblock Array detection for speckle reduction in optical coherence
  microscopy.
\newblock {\em Phys Med Biol}, 42(7):1427--39, 1997.

\bibitem{RN154}
J.~M. Schmitt.
\newblock Restoration of optical coherence images of living tissue using the
  clean algorithm.
\newblock {\em J Biomed Opt}, 3(1):66--75, 1998.

\bibitem{RN151}
J.~M. Schmitt, S.~H. Xiang, and K.~M. Yung.
\newblock Speckle in optical coherence tomography.
\newblock {\em J Biomed Opt}, 4(1):95--105, 1999.

\bibitem{RN100}
V.~Behar, D.~Adam, and Z.~Friedman.
\newblock A new method of spatial compounding imaging.
\newblock {\em Ultrasonics}, 41(5):377--84, 2003.

\bibitem{RN10}
Wei Wu, Ou~Tan, Rajeev~R. Pappuru, Huilong Duan, and David Huang.
\newblock Assessment of frame-averaging algorithms in oct image analysis.
\newblock {\em Ophthalmic surgery, lasers \& imaging retina}, 44(2):168--175,
  2013.

\bibitem{RN11}
Chieh-Li Chen, Hiroshi Ishikawa, Gadi Wollstein, Richard~A. Bilonick, Larry
  Kagemann, and Joel~S. Schuman.
\newblock Virtual averaging making nonframe-averaged optical coherence
  tomography images comparable to frame-averaged images.
\newblock {\em Translational Vision Science \& Technology}, 5(1):1, 2016.

\bibitem{RN1}
Shahab Chitchian, Markus~A. Mayer, Adam~R. Boretsky, Frederik~J. van Kuijk, and
  Massoud Motamedi.
\newblock Retinal optical coherence tomography image enhancement via shrinkage
  denoising using double-density dual-tree complex wavelet transform.
\newblock {\em Journal of Biomedical Optics}, 17(11):116009, 2012.

\bibitem{RN3}
R.~D. Ferguson, D.~X. Hammer, L.~A. Paunescu, S.~Beaton, and J.~S. Schuman.
\newblock Tracking optical coherence tomography.
\newblock {\em Opt Lett}, 29(18):2139--41, 2004.

\bibitem{RN161}
A.~Ozcan, A.~Bilenca, A.~E. Desjardins, B.~E. Bouma, and G.~J. Tearney.
\newblock Speckle reduction in optical coherence tomography images using
  digital filtering.
\newblock {\em J Opt Soc Am A Opt Image Sci Vis}, 24(7):1901--10, 2007.

\bibitem{RN162}
R.~Bernardes, C.~Maduro, P.~Serranho, A.~Araujo, S.~Barbeiro, and J.~Cunha-Vaz.
\newblock Improved adaptive complex diffusion despeckling filter.
\newblock {\em Opt Express}, 18(23):24048--59, 2010.

\bibitem{RN160}
A.~Wong, A.~Mishra, K.~Bizheva, and D.~A. Clausi.
\newblock General bayesian estimation for speckle noise reduction in optical
  coherence tomography retinal imagery.
\newblock {\em Opt Express}, 18(8):8338--52, 2010.

\bibitem{RN37}
Feng~Chen Liheng~Bian, Jinli~Suo and Qionghai Dai.
\newblock Multi-frame denoising of high speed optical coherence tomography data
  using inter-frame and intra-frame priors.
\newblock {\em arXiv:1312.1931 [cs.CV]}, 2014.

\bibitem{RN31}
N.~M. Grzywacz, J.~de~Juan, C.~Ferrone, D.~Giannini, D.~Huang, G.~Koch,
  V.~Russo, O.~Tan, and C.~Bruni.
\newblock Statistics of optical coherence tomography data from human retina.
\newblock {\em IEEE Trans Med Imaging}, 29(6):1224--37, 2010.

\bibitem{RN30}
Milan~Sonka Hossein~Rabbani and Michael~D. Abramoff.
\newblock Optical coherence tomography noise reduction using anisotropic local
  bivariate gaussian mixture prior in 3d complex wavelet domain.
\newblock {\em International Journal of Biomedical Imaging}, 2013, 2013.

\bibitem{RN64}
Muxingzi Li, Ramzi Idoughi, Biswarup Choudhury, and Wolfgang Heidrich.
\newblock Statistical model for oct image denoising.
\newblock {\em Biomedical Optics Express}, 8(9):3903--3917, 2017.

\bibitem{RN35}
Markus~A. Mayer, Anja Borsdorf, Martin Wagner, Joachim Hornegger, Christian~Y.
  Mardin, and Ralf~P. Tornow.
\newblock Wavelet denoising of multiframe optical coherence tomography data.
\newblock {\em Biomedical Optics Express}, 3(3):572--589, 2012.

\bibitem{RN61}
I.~Y. Wong, H.~Koizumi, and W.~W. Lai.
\newblock Enhanced depth imaging optical coherence tomography.
\newblock {\em Ophthalmic Surg Lasers Imaging}, 42 Suppl:S75--84, 2011.

\bibitem{RN12}
Lelia Adelina Paunescu Siobahn Beaton Joel S.~Schuman R.~Daniel~Ferguson,
  Daniel X.~Hammer.
\newblock Tracking optical coherence tomography.
\newblock {\em Optics Letters}, 29(18), 2004.

\bibitem{RN70}
D.~Hammer, R.~D. Ferguson, N.~Iftimia, T.~Ustun, G.~Wollstein, H.~Ishikawa,
  M.~Gabriele, W.~Dilworth, L.~Kagemann, and J.~Schuman.
\newblock Advanced scanning methods with tracking optical coherence tomography.
\newblock {\em Opt Express}, 13(20):7937--47, 2005.

\bibitem{RN72}
3rd Wells, W.~M., P.~Viola, H.~Atsumi, S.~Nakajima, and R.~Kikinis.
\newblock Multi-modal volume registration by maximization of mutual
  information.
\newblock {\em Med Image Anal}, 1(1):35--51, 1996.

\bibitem{RN107}
Rupal~R. Agravat and Mehul~S. Raval.
\newblock {\em Chapter 11 - Deep Learning for Automated Brain Tumor
  Segmentation in MRI Images}, pages 183--201.
\newblock Academic Press, 2018.

\bibitem{RN104}
Zeynettin Akkus, Alfiia Galimzianova, Assaf Hoogi, Daniel~L. Rubin, and
  Bradley~J. Erickson.
\newblock Deep learning for brain mri segmentation: State of the art and future
  directions.
\newblock {\em Journal of Digital Imaging}, 30(4):449--459, 2017.

\bibitem{RN106}
Z.~Cui, J.~Yang, and Y.~Qiao.
\newblock Brain mri segmentation with patch-based cnn approach.
\newblock In {\em 2016 35th Chinese Control Conference (CCC)}, pages
  7026--7031.

\bibitem{RN105}
J.~Liu, Y.~Pan, M.~Li, Z.~Chen, L.~Tang, C.~Lu, and J.~Wang.
\newblock Applications of deep learning to mri images: A survey.
\newblock {\em Big Data Mining and Analytics}, 1(1):1--18, 2018.

\bibitem{RN108}
Heba Mohsen, El-Sayed~A. El-Dahshan, El-Sayed~M. El-Horbaty, and
  Abdel-Badeeh~M. Salem.
\newblock Classification using deep learning neural networks for brain tumors.
\newblock {\em Future Computing and Informatics Journal}, 3(1):68--71, 2018.

\bibitem{RN109}
Joachim~Buhmann Viktor~Wegmayr, Sai~Aitharaju.
\newblock Classification of brain mri with big data and deep 3d convolutional
  neural networks.
\newblock {\em Proceedings Volume 10575, Medical Imaging 2018: Computer-Aided
  Diagnosis; 105751S}, 2018.

\bibitem{RN110}
Roberto Ardon Isabelle~Bloch Hadrien~Bertrand, Matthieu~Perrot.
\newblock Classification of mri data using deep learning and gaussian
  process-based model selection.
\newblock {\em arXiv:1701.04355 [cs.LG]}, 2017.

\bibitem{RN113}
A.~Benou, R.~Veksler, A.~Friedman, and T.~Riklin~Raviv.
\newblock Ensemble of expert deep neural networks for spatio-temporal denoising
  of contrast-enhanced mri sequences.
\newblock {\em Med Image Anal}, 42:145--159, 2017.

\bibitem{RN111}
Luc Vosters Xiayu Xu Yue Sun Tao~Tan Dongsheng~Jiang, Weiqiang~Dou.
\newblock Denoising of 3d magnetic resonance images with multi-channel residual
  learning of convolutional neural network.
\newblock {\em arXiv:1712.08726 [cs.CV]}, 2017.

\bibitem{RN112}
Lovedeep Gondara.
\newblock Medical image denoising using convolutional denoising autoencod.
\newblock {\em arXiv:1608.04667v2 [cs.CV]}, 2016.

\bibitem{RN120}
Xiaodan Sui, Yuanjie Zheng, Benzheng Wei, Hongsheng Bi, Jianfeng Wu, Xuemei
  Pan, Yilong Yin, and Shaoting Zhang.
\newblock Choroid segmentation from optical coherence tomography with
  graph-edge weights learned from deep convolutional neural networks.
\newblock {\em Neurocomputing}, 237:332--341, 2017.

\bibitem{RN119}
B.~Al-Bander, B.~M. Williams, M.~A. Al-Taee, W.~Al-Nuaimy, and Y.~Zheng.
\newblock A novel choroid segmentation method for retinal diagnosis using deep
  learning.
\newblock In {\em 2017 10th International Conference on Developments in
  eSystems Engineering (DeSE)}, pages 182--187.

\bibitem{RN118}
Qiao Zhang, Zhipeng Cui, Xiaoguang Niu, Shijie Geng, and Yu~Qiao.
\newblock Image segmentation with pyramid dilated convolution based on resnet
  and u-net.
\newblock In Derong Liu, Shengli Xie, Yuanqing Li, Dongbin Zhao, and
  El-Sayed~M. El-Alfy, editors, {\em Neural Information Processing}, pages
  364--372. Springer International Publishing.

\bibitem{RN117}
Freerk~G. Venhuizen, Bram van Ginneken, Bart Liefers, Mark J. J.~P. van
  Grinsven, Sascha Fauser, Carel Hoyng, Thomas Theelen, and Clara~I. Sánchez.
\newblock Robust total retina thickness segmentation in optical coherence
  tomography images using convolutional neural networks.
\newblock {\em Biomedical Optics Express}, 8(7):3292--3316, 2017.

\bibitem{RN116}
Leyuan Fang, David Cunefare, Chong Wang, Robyn~H. Guymer, Shutao Li, and Sina
  Farsiu.
\newblock Automatic segmentation of nine retinal layer boundaries in oct images
  of non-exudative amd patients using deep learning and graph search.
\newblock {\em Biomedical Optics Express}, 8(5):2732--2744, 2017.

\bibitem{RN115}
Sieun Lee Gavin Ding Marinko~V.Sarunic Donghuan~Lu, Morgan~Heisler and
  Mirza~Faisal Beg.
\newblock Retinal fluid segmentation and detection in optical coherence
  tomography images using fully convolutional neural network.
\newblock {\em arXiv:1710.04778v1 [cs.CV] 13 Oct 2017}, 2017.

\bibitem{RN84}
Sri Phani Krishna Karri Debdoot Sheet Amin Katouzian Christian Wachinger
  Nassir~Navab Abhijit Guha~Roy, Sailesh~Conjeti.
\newblock Relaynet: Retinal layer and fluid segmentation of macular optical
  coherence tomography using fully convolutional networks.
\newblock {\em arXiv:1704.02161v2 [cs.CV] 7 Jul 2017}.

\bibitem{RN103}
Sripad~Krishna Devalla, Prajwal~K. Renukanand, Bharathwaj~K. Sreedhar, Giridhar
  Subramanian, Liang Zhang, Shamira Perera, Jean-Martial Mari, Khai~Sing Chin,
  Tin~A. Tun, Nicholas~G. Strouthidis, Tin Aung, Alexandre~H. Thiéry, and
  Michaël J.~A. Girard.
\newblock Drunet: a dilated-residual u-net deep learning network to segment
  optic nerve head tissues in optical coherence tomography images.
\newblock {\em Biomedical Optics Express}, 9(7):3244--3265, 2018.

\bibitem{RN82}
S.~K. Devalla, K.~S. Chin, J.~M. Mari, T.~A. Tun, N.~G. Strouthidis, T.~Aung,
  A.~H. Thiery, and M.~J.~A. Girard.
\newblock A deep learning approach to digitally stain optical coherence
  tomography images of the optic nerve head.
\newblock {\em Invest Ophthalmol Vis Sci}, 59(1):63--74, 2018.

\bibitem{RN123}
M.~Awais, H.~Müller, T.~B. Tang, and F.~Meriaudeau.
\newblock Classification of sd-oct images using a deep learning approach.
\newblock In {\em 2017 IEEE International Conference on Signal and Image
  Processing Applications (ICSIPA)}, pages 489--492.

\bibitem{RN122}
Philipp Prahs, Viola Radeck, Christian Mayer, Yordan Cvetkov, Nadezhda
  Cvetkova, Horst Helbig, and David Märker.
\newblock Oct-based deep learning algorithm for the evaluation of treatment
  indication with anti-vascular endothelial growth factor medications.
\newblock {\em Graefe's Archive for Clinical and Experimental Ophthalmology},
  256(1):91--98, 2018.

\bibitem{RN121}
Cecilia~S. Lee, Doug~M. Baughman, and Aaron~Y. Lee.
\newblock Deep learning is effective for classifying normal versus age-related
  macular degeneration oct images.
\newblock {\em Ophthalmology Retina}, 1(4):322--327, 2017.

\bibitem{RN74}
O.~Ronneberger Brox, P.Fischer, and T.
\newblock U-net: Convolutional networks for biomedical image segmentation.
\newblock {\em Medical Image Computing and Computer-Assisted Intervention
  (MICCAI)}, 9351:234--241, 2015.

\bibitem{RN76}
Kaiming~He Sun, Xiangyu Zhang, Shaoqing Ren, and Jian.
\newblock Deep residual learning for image recognition.
\newblock {\em arXiv preprint arXiv:1512.03385}, 2015.

\bibitem{RN77}
Fisher~Yu Koltun and Vladlen.
\newblock Multi-scale context aggregation by dilated convolutions.
\newblock {\em ICLR}.

\bibitem{RN90}
Y.~Liu, M.~M. Cheng, X.~Hu, K.~Wang, and X.~Bai.
\newblock Richer convolutional features for edge detection.
\newblock In {\em 2017 IEEE Conference on Computer Vision and Pattern
  Recognition (CVPR)}, pages 5872--5881.

\bibitem{RN91}
Djork{-}Arn{\'{e}}~Clevert and, Thomas~Unterthiner and, and Sepp Hochreiter.
\newblock Fast and accurate deep network learning by exponential linear
  units(elus).
\newblock {\em CoRR}, abs/1511.07289, 2015.

\bibitem{RN92}
Christian~Szegedy Sergey~Ioffe.
\newblock Batch normalization: accelerating deep network training by reducing
  internal covariate shift.
\newblock In {\em Proceedings of the 32nd International Conference on
  International Conference on Machine Learning - Volume 37}, pages 448--456.

\bibitem{RN93}
Jimmy~Ba Diederik P.~Kingma.
\newblock Adam: A method for stochastic optimization.
\newblock {\em arXiv:1412.6980}, 2014.

\bibitem{RN94}
John C.~Platt Patrice Y.~Simard, Dave~Steinkraus.
\newblock Best practices for convolutional neural networks applied to visual
  document analysis.
\newblock {\em Proceedings of the Seventh International Conference on Document
  Analysis and Recognition (ICDAR 2003)}, 2003.

\bibitem{RN95}
Z.~Wu, G.~Xu, R.~N. Weinreb, M.~Yu, and C.~K. Leung.
\newblock Optic nerve head deformation in glaucoma: A prospective analysis of
  optic nerve head surface and lamina cribrosa surface displacement.
\newblock {\em Ophthalmology}, 122(7):1317--29, 2015.

\bibitem{RN163}
M.~J. Girard, N.~G. Strouthidis, C.~R. Ethier, and J.~M. Mari.
\newblock Shadow removal and contrast enhancement in optical coherence
  tomography images of the human optic nerve head.
\newblock {\em Invest Ophthalmol Vis Sci}, 52(10):7738--48, 2011.

\bibitem{RN166}
Wang Zhou, A.~C. Bovik, H.~R. Sheikh, and E.~P. Simoncelli.
\newblock Image quality assessment: from error visibility to structural
  similarity.
\newblock {\em IEEE Transactions on Image Processing}, 13(4):600--612, 2004.

\bibitem{RN130}
Abdullah Al-Mujaini, Upender~K. Wali, and Sitara Azeem.
\newblock Optical coherence tomography: Clinical applications in medical
  practice.
\newblock {\em Oman Medical Journal}, 28(2):86--91, 2013.

\bibitem{RN45}
C.~Bowd, R.~N. Weinreb, J.~M. Williams, and L.~M. Zangwill.
\newblock The retinal nerve fiber layer thickness in ocular hypertensive,
  normal, and glaucomatous eyes with optical coherence tomography.
\newblock {\em Arch Ophthalmol}, 118(1):22--6, 2000.

\bibitem{RN129}
Gillian~J. McLellan and Carol~A. Rasmussen.
\newblock Optical coherence tomography for the evaluation of retinal and optic
  nerve morphology in animal subjects: Practical considerations.
\newblock {\em Veterinary ophthalmology}, 15(Suppl 2):13--28, 2012.

\bibitem{RN46}
A.~Miki, F.~A. Medeiros, R.~N. Weinreb, S.~Jain, F.~He, L.~Sharpsten,
  N.~Khachatryan, N.~Hammel, J.~M. Liebmann, C.~A. Girkin, P.~A. Sample, and
  L.~M. Zangwill.
\newblock Rates of retinal nerve fiber layer thinning in glaucoma suspect eyes.
\newblock {\em Ophthalmology}, 121(7):1350--8, 2014.

\bibitem{RN47}
T.~Ojima, T.~Tanabe, M.~Hangai, S.~Yu, S.~Morishita, and N.~Yoshimura.
\newblock Measurement of retinal nerve fiber layer thickness and macular volume
  for glaucoma detection using optical coherence tomography.
\newblock {\em Jpn J Ophthalmol}, 51(3):197--203, 2007.

\bibitem{RN56}
J.~C. Downs, M.~E. Ensor, A.~J. Bellezza, H.~W. Thompson, R.~T. Hart, and C.~F.
  Burgoyne.
\newblock Posterior scleral thickness in perfusion-fixed normal and
  early-glaucoma monkey eyes.
\newblock {\em Invest Ophthalmol Vis Sci}, 42(13):3202--8, 2001.

\bibitem{RN57}
K.~M. Lee, T.~W. Kim, R.~N. Weinreb, E.~J. Lee, M.~J. Girard, and J.~M. Mari.
\newblock Anterior lamina cribrosa insertion in primary open-angle glaucoma
  patients and healthy subjects.
\newblock {\em PLoS One}, 9(12):e114935, 2014.

\bibitem{RN52}
S.~C. Park, J.~Brumm, R.~L. Furlanetto, C.~Netto, Y.~Liu, C.~Tello, J.~M.
  Liebmann, and R.~Ritch.
\newblock Lamina cribrosa depth in different stages of glaucoma.
\newblock {\em Invest Ophthalmol Vis Sci}, 56(3):2059--64, 2015.

\bibitem{RN54}
H.~A. Quigley and E.~M. Addicks.
\newblock Regional differences in the structure of the lamina cribrosa and
  their relation to glaucomatous optic nerve damage.
\newblock {\em Arch Ophthalmol}, 99(1):137--43, 1981.

\bibitem{RN53}
H.~A. Quigley, E.~M. Addicks, W.~R. Green, and A.~E. Maumenee.
\newblock Optic nerve damage in human glaucoma. ii. the site of injury and
  susceptibility to damage.
\newblock {\em Arch Ophthalmol}, 99(4):635--49, 1981.

\bibitem{RN127}
Caio~V. Regatieri, Lauren Branchini, and Jay~S. Duker.
\newblock The role of spectral-domain oct in the diagnosis and management of
  neovascular age-related macular degeneration.
\newblock {\em Ophthalmic surgery, lasers \& imaging : the official journal of
  the International Society for Imaging in the Eye}, 42(0):S56--S66, 2011.

\bibitem{RN131}
Ontario Health~Quality.
\newblock Optical coherence tomography for age-related macular degeneration and
  diabetic macular edema: An evidence-based analysis.
\newblock {\em Ontario Health Technology Assessment Series}, 9(13):1--22, 2009.

\bibitem{RN132}
Vivek~J. Srinivasan, Maciej Wojtkowski, Andre~J. Witkin, Jay~S. Duker, Tony~H.
  Ko, Mariana Carvalho, Joel~S. Schuman, Andrzej Kowalczyk, and James~G.
  Fujimoto.
\newblock High-definition and 3-dimensional imaging of macular pathologies with
  high-speed ultrahigh-resolution optical coherence tomography.
\newblock {\em Ophthalmology}, 113(11):2054.e1--2054.14, 2006.

\bibitem{RN22}
Rayan~A. Alshareef, Sunila Dumpala, Shruthi Rapole, Manideepak Januwada,
  Abhilash Goud, Hari~Kumar Peguda, and Jay Chhablani.
\newblock Prevalence and distribution of segmentation errors in macular
  ganglion cell analysis of healthy eyes using cirrus hd-oct.
\newblock {\em PLOS ONE}, 11(5):e0155319, 2016.

\bibitem{RN288}
A.~Stankiewicz, T.~Marciniak, A.~Dąbrowski, M.~Stopa, P.~Rakowicz, and
  E.~Marciniak.
\newblock Denoising methods for improving automatic segmentation in oct images
  of human eye.
\newblock {\em Bulletin of the Polish Academy of Sciences Technical Sciences},
  65(1), 2017.

\bibitem{RN136}
Drew Scoles, John~A. Flatter, Robert~F. Cooper, Christopher~S. Langlo, Scott
  Robison, Maureen Neitz, David~V. Weinberg, Mark~E. Pennesi, Dennis~P. Han,
  Alfredo Dubra, and Joseph Carroll.
\newblock Assessing photoreceptor structure associated with ellipsoid zone
  disruptions visualized with optical coherence tomography.
\newblock {\em Retina (Philadelphia, Pa.)}, 36(1):91--103, 2016.

\bibitem{RN139}
Timothy~S. Kern and Bruce~A. Berkowitz.
\newblock Photoreceptors in diabetic retinopathy.
\newblock {\em Journal of Diabetes Investigation}, 6(4):371--380, 2015.

\bibitem{RN137}
T.~Baba, S.~Yamamoto, M.~Arai, E.~Arai, T.~Sugawara, Y.~Mitamura, and
  S.~Mizunoya.
\newblock Correlation of visual recovery and presence of photoreceptor
  inner/outer segment junction in optical coherence images after successful
  macular hole repair.
\newblock {\em Retina}, 28(3):453--8, 2008.

\bibitem{RN138}
Hisako Hayashi, Kenji Yamashiro, Akitaka Tsujikawa, Masafumi Ota, Atsushi
  Otani, and Nagahisa Yoshimura.
\newblock Association between foveal photoreceptor integrity and visual outcome
  in neovascular age-related macular degeneration.
\newblock {\em American Journal of Ophthalmology}, 148(1):83--89.e1, 2009.

\bibitem{RN140}
J.~A. Flatter, R.~F. Cooper, M.~J. Dubow, A.~Pinhas, R.~S. Singh, R.~Kapur,
  N.~Shah, R.~D. Walsh, S.~H. Hong, D.~V. Weinberg, K.~E. Stepien, W.~J.
  Wirostko, S.~Robison, A.~Dubra, R.~B. Rosen, Jr. Connor, T.~B., and
  J.~Carroll.
\newblock Outer retinal structure after closed-globe blunt ocular trauma.
\newblock {\em Retina}, 34(10):2133--46, 2014.

\bibitem{RN143}
Maria~P. Bambo, Elena Garcia-Martin, Sofia Otin, Eva Sancho, Isabel Fuertes,
  Raquel Herrero, Maria Satue, and Luis Pablo.
\newblock Influence of cataract surgery on repeatability and measurements of
  spectral domain optical coherence tomography.
\newblock {\em British Journal of Ophthalmology}, 98(1):52, 2014.

\bibitem{RN144}
Pauline H.~B. Kok, Thomas J. T.~P. van~den Berg, Hille~W. van Dijk, Marilette
  Stehouwer, Ivanka J.~E. van~der Meulen, Maarten~P. Mourits, and Frank~D.
  Verbraak.
\newblock The relationship between the optical density of cataract and its
  influence on retinal nerve fibre layer thickness measured with spectral
  domain optical coherence tomography.
\newblock {\em Acta Ophthalmologica}, 91(5):418--424, 2012.

\bibitem{RN142}
J.~C. Mwanza, A.~M. Bhorade, N.~Sekhon, J.~J. McSoley, S.~H. Yoo, W.~J. Feuer,
  and D.~L. Budenz.
\newblock Effect of cataract and its removal on signal strength and
  peripapillary retinal nerve fiber layer optical coherence tomography
  measurements.
\newblock {\em J Glaucoma}, 20(1):37--43, 2011.

\bibitem{RN145}
G.~Savini, M.~Zanini, and P.~Barboni.
\newblock Influence of pupil size and cataract on retinal nerve fiber layer
  thickness measurements by stratus oct.
\newblock {\em J Glaucoma}, 15(4):336--40, 2006.

\bibitem{RN141}
Daniel~M. Stein, Gadi Wollstein, Hiroshi Ishikawa, Ellen Hertzmark, Robert~J.
  Noecker, and Joel~S. Schuman.
\newblock Effect of corneal drying on optical coherence tomography.
\newblock {\em Ophthalmology}, 113(6):985--991, 2006.

\bibitem{RN146}
Jason W.~Much Nicola G.~Ghazi.
\newblock The effect of lubricating eye drops on optical coherence tomography
  imaging of the retina.
\newblock {\em Digital Journal of Ophthalmology}, 15(2), 2009.

\bibitem{RN147}
Michael Smith, Andrew Frost, Christopher~Mark Graham, and Steven Shaw.
\newblock Effect of pupillary dilatation on glaucoma assessments using optical
  coherence tomography.
\newblock {\em The British Journal of Ophthalmology}, 91(12):1686--1690, 2007.

\bibitem{RN148}
E.~Moisseiev, D.~Loberman, E.~Zunz, A.~Kesler, A.~Loewenstein, and
  J.~Mandelblum.
\newblock Pupil dilation using drops vs gel: a comparative study.
\newblock {\em Eye}, 29:815, 2015.

\bibitem{RN165}
Joong~Won Shin, Yong~Un Shin, Ki~Bang Uhm, Kyung~Rim Sung, Min~Ho Kang,
  Hee~Yoon Cho, and Mincheol Seong.
\newblock The effect of optic disc center displacement on retinal nerve fiber
  layer measurement determined by spectral domain optical coherence tomography.
\newblock {\em PLOS ONE}, 11(10):e0165538, 2016.

\end{thebibliography}

\end{document}